# Towards a Framework for Visual Intelligence in Service Robotics: Epistemic Requirements and Gap Analysis

**Agnese Chiatti, Enrico Motta, Enrico Daga**
Knowledge Media Institute, The Open University, United Kingdom
{agnese.chiatti, enrico.motta, enrico.daga}@open.ac.uk

## Abstract

A key capability required by service robots operating in real-world, dynamic environments is that of *Visual Intelligence*, i.e., the ability to use their vision system, reasoning components and background knowledge to make sense of their environment. In this paper, we analyze the epistemic requirements for Visual Intelligence, both in a top-down fashion, using existing frameworks for human-like Visual Intelligence in the literature, and from the bottom up, based on the errors emerging from object recognition trials in a real-world robotic scenario. Finally, we use these requirements to evaluate current knowledge bases for Service Robotics and to identify gaps in the support they provide for Visual Intelligence. These gaps provide the basis of a research agenda for developing more effective knowledge representations for Visual Intelligence.

## 1. Introduction

The fast-paced advancement of the Artificial Intelligence (AI) and Robotics fields has drastically reduced the technological and economic barriers traditionally ascribed to the design of real-world robotic applications. Thanks to these developments, there is an increased potential for designing and deploying robots that can assist people with their daily tasks, i.e., *service robots*. The possible range of services is vast: from Health and Safety monitoring (Bastianelli *et al.*, 2018; Dong *et al.*, 2018), to pre-emptive patient care (Mollaret *et al.*, 2018; Bajones et al., 2018), door-to-door garbage collection (Ferri *et al.*, 2011), and others.

A critical capability required by service robots operating in real-world, dynamic environments is that of *Visual Intelligence*, i.e., the ability to use their vision system, reasoning components and external knowledge sources to make sense of their environment. Let us consider the case of HanS, the Health and Safety (H&S) robot inspector under development at the Knowledge Media Institute (KMi) (Bastianelli *et al.*, 2018). HanS is expected to monitor the Lab space in search of potentially harming situations, such as a fire hazard. To interpret the state of the environment (e.g., identify the presence of a fire hazard) and act upon it (e.g., notify the designated fire wardens), HanS needs to recognize the objects in each image captured from its camera sensor, and correctly reason about the content of each image. For instance, to recognize the risk posed by a *portable heater* sitting on top of a *pile of paper*, HanS would need to recognize not only that (i) a *heater* and a *pile of paper* are there, but also that (ii) these two objects are *close to* each other, (iii) that *portable heaters*, like other *electric devices*, can produce *heat* (iv) that *paper is more likely to catch fire* than other materials, and (v) that *the proximity of ignition sources to flammable materials is a fire hazard*.

Even from this simple example, it is evident that, to fulfill its assistance duties, a service robot needs not only object recognition but also broader sensemaking capabilities. In particular, in this paper we focus on the *Visual Intelligence* of a robot, as a prerequisite for sensemaking[1].

To better pinpoint *the set of capabilities and knowledge properties required for service robots to exhibit Visual Intelligence*, we start from related research on Machine Intelligence (Lake *et al.*, 2017) and Visual Cognition (Hoffman, 2000), in a top-down fashion. Then, we qualitatively analyze the requirements emerging from the object recognition performance achieved by HanS during our trials (Chiatti *et al.*, 2019). Finally, we discuss the extent to which these bottom-up requirements align with the requirements derived from the top down.

Considering the current limitations of state-of-the-art Image Understanding methods, which are purely based on Machine Learning (ML), we also identify a set of knowledge bases which can augment the existing solutions. These include (i) knowledge representations explicitly conceived for robotic applications, (ii) other general-purpose knowledge sources which can still be of help to a service robot, due to their scale, and (iii) the benchmark datasets in Image Understanding. We then use the knowledge requirements identified in the previous tasks to evaluate *to which extent the selected Knowledge Bases can effectively support Visual Intelligence in Service Robotics*.

---

[1] Needless to say, other sensory modalities can also contribute to the sensemaking capabilities of robot assistants. For instance, sound can be used to detect noise levels which may put an employee's health at risk.

## 2. Background and motivation

### 2.1. Computer Vision and Image Understanding

The first prerequisite for a service robot like HanS to attend to its tasks is understanding the content of its observations, i.e., Image Understanding. The human-like, or even above-human (Krizhevsky *et al.*, 2012) performance which Deep Learning based methods have shown on several benchmarks (Redmon and Farhadi, 2018; Ren *et al.*, 2015) has produced much excitement in the field of Computer Vision. As a result, Deep Neural Networks (NNs) have become the *de facto* methodology for most Image Understanding tasks.

However, Deep Neural Networks are (i) notoriously data-hungry, (ii) largely based on learning offline with respect to a set of pre-determined categories, i.e., work under the *closed world assumption* (Mancini *et al.*, 2019), and (iii) prone to *catastrophically forgetting* previously learned concepts, once new concepts are introduced (Parisi *et al.*, 2019).

Moreover, Deep Learning is based on data representations derived indirectly, by backpropagating through thousands of training exemplars, rather than explicitly, from feature engineering. The latter trait is a double-edged sword. On the one hand, it removes the startup costs and burden of modelling a new application scenario explicitly. On the other hand, it makes tasks such as explaining the obtained features, reasoning about world states, and integrating explicit knowledge statements far from trivial (Marcus, 2018).

Deep NNs have exhibited impressive performance on certain Computer Vision tasks. However, machine Visual Intelligence is still inferior to human Visual Intelligence in many ways (Lake *et al.*, 2017). Humans can learn to recognize new object categories almost instantly, from just a few observations. While Deep NNs are designed to recognize patterns from the input data, humans can learn richer object representations even from minimal and sparse observations, forming a mental "blueprint of their environment" (Pearl, 2018), a process also referred to as *model building* by Lake *et al.* (2017). In constructing their visual world, they can overcome the most fundamental vision problem: that each retinal image has countless possible interpretations in the 3D world (Hoffman, 2000).

As Lake *et al.* (2017) emphasize, this evidence is not presented against the use of Deep NNs. Deep Learning methods can provide a useful baseline to bootstrap an object recognition system and ensure near-real-time recognition speed on known object classes, i.e., classes seen at training time. However, purely Machine Learning (ML) based methods need to be complemented by other, richer knowledge representations to equip service robots with mechanisms to adapt to uncertainty and learn new objects and concepts over time. This awareness has recently led to the development of Image Understanding systems which integrate external knowledge with Deep NNs. A detailed survey of these methods can be found in (Aditya *et al.*, 2019) and in (Gouidis *et al.*, 2019). Both reviews discuss the advantages and limitations of different hybrid (i.e., both ML-based and knowledge-based) learning models. However, the question of *which types of knowledge representations to leverage within hybrid learning models* remains open (Daruna *et al.*, 2018). The first step is thus to discuss the state of the art in Knowledge Representation for Service Robots.

### 2.2. Knowledge Representation for Service Robots

Following Paulius and Sun's definition (2019), a suitable knowledge representation should bridge the gap between the lower-level inputs collected by the robot's perceptual layers (e.g., through vision and navigation) and the higher-level, semantic representation of these symbols. Paulius and Sun discriminate between *specific* and *comprehensive (or fully-fledged) knowledge representations*. Learning models produce specific representations (e.g., the image embeddings in a Convolutional Neural Network's layers; the directed, acyclic graphs in a Bayesian Network). Comprehensive knowledge bases, instead, formalize relevant concepts as higher-level ontologies and are agnostic to the specific learning method used. Considering the breadth of knowledge required for intelligent systems to exhibit commonsense (Davis and Marcus, 2015) and human-like Visual Intelligence (Lake *et al.*, 2017; Hoffman, 2000), in what follows, we prioritize the analysis of knowledge representations which can be considered comprehensive. Naturally, these fully-fledged KBs could then augment other lower-level representations, e.g., the image embeddings of the learning methods discussed in the previous section, or the geometric maps depicting the robot's environment, e.g., leading to enhanced, semantic maps (Nüchter and Hertzberg, 2008).

**Resources designed specifically for Robots. KnowRob** (Tenorth and Beetz, 2009; Beetz *et al.*, 2018) is, to date, the most comprehensive knowledge base for robots (Paulius and Sun, 2019; Thosar *et al.*, 2018). Made partially available through the Open-EASE platform (Beetz *et al.*, 2015), KnowRob integrates: (i) a core ontology built on top of **OpenCyc** (Lenat, 1995), (ii) web-mined data including encyclopaedic web pages (i.e., how-tos and tutorials), recipe databases, specific online shops, and (iii) semantically-annotated observations of human demonstrations. Concepts and relations in KnowRob are defined through propositional logic. Within this knowledge processing pipeline, perception is handled through the RoboSherlock Vision suite (Beetz *et al.*, 2015). The robot's observations are then validated manually (Bálint-Benczédi and Beetz, 2018), to be consolidated in the form of *episodic memories* (Bálint-Benczédi *et al.*, 2017). A photo-realistic rendering of the robot's environment is used to simulate alternative memories as well as predict the outcome of certain actions, through a physics game engine (Beetz *et al.*, 2018).

**General-purpose Resources.** Besides the KBs explicitly designed for the robotic domain, many other large-scale knowledge sources are available. In a recent survey, Storks and colleagues (2019) have categorized these resources as

*linguistic*, *common* and *commonsense knowledge*, based on the type of properties they encode.

Linguistic knowledge provides tools to understand "the word meanings, grammar, syntax, semantics and discourse structure" (Storks et al., 2019). **WordNet** (Miller, 1995) is the most extensive word lexicon in English, where synonym words are grouped in *synsets* and linked to their hypernyms, hyponyms, antonyms and entailed concepts. Another linguistic reference is the **Unified Verb Index[2] (UVI)**. Conveniently, UVI merges the verb groupings of four core verb repositories, namely **VerbNet** (Schuler, 2005), **FrameNet** (Fillmore *et al.*, 2003), **PropBank** (Kingsbury and Palmer, 2002), and the sense groupings resulting from the **OntoNotes** annotation initiative (Hovy *et al.*, 2006).

It is essential to differentiate between *common knowledge*, comprising of "known facts about the world which can be explicitly stated" (Storks *et al.*, 2019) and *commonsense knowledge*, which is typically taken for granted by humans and is, therefore, harder to formalize (Davis and Marcus, 2015).

Large-scale collections of factual knowledge can be derived from Wikipedia articles and infoboxes, as in the case of **YAGO** (Suchanek *et al.*, 2007), **DBpedia** (Auer *et al.*, 2007) and **Wikidata** (Vrandečić and Krötzsch, 2014). As a result, the content of these encyclopaedic sources partially overlaps. However, Wikidata also includes concepts gradually migrated from **FreeBase** (Bollacker *et al.*, 2008), a collaboratively created repository of facts officially decommissioned in 2014. **Probase** (Wu *et al.*, 2012) and **NELL** (Carlson *et al.*, 2010), instead, are collections of facts mined from a broader set of web pages. Probase is currently exposed as part of the Microsoft Concept Graph. Beliefs in NELL have been mined incrementally since 2010.

Attempts have been made to infer commonsense knowledge from everyday facts, as in the case of **ConceptNet** (Liu and Singh, 2004). ConceptNet consists of common and commonsense statements collected from online users, augmented with concepts derived from OpenCyc, WordNet and DBpedia (Speer *et al.*, 2017). While the core of ConceptNet is the result of a crowd-sourced effort, **WebChild** (Tandon *et al.*, 2017) includes noun-adjective commonsense relations automatically mined from the Web. **ATOMIC** (Sap *et al.*, 2019) and **ASER** (Zhang *et al.*, 2019) are extensive collections of inferential knowledge represented as "if-then" triplets of everyday events.

**Resources specific to Image Understanding.** In the context of Image Understanding, another key aspect is linking the linguistic, common and commonsense textual sources discussed in the previous Sections with imagery. A set of relevant KBs for Image Understanding can be derived from (Wu *et al.*, 2017) and (Liu *et al.*, 2019). Here we focus on the image collections, among those identified in the last two surveys, which have been mapped to the taxonomies discussed in the previous sections, to facilitate the linking of different knowledge sources.

**Visual Genome (VG)** (Krishna *et al.*, 2017) includes natural images from the intersection of YFCC100M (Thomee *et al.*, 2016) and MS-COCO (Lin *et al.*, 2014). Scenes are annotated with regions enclosing each object. Each region is annotated with: (i) the object class label, (ii) a textual description of the region content, and, optionally, (iii) additional object attributes (e.g., colour, state, and others). Moreover, VG also includes, for each image: (iv) the object-object relationships connecting different object regions, i.e., a *scene graph*, and (v) a set of sample Q&A about the scene.

Crucially, Wu *et al.* (2017) found that only 40.02% of the correct answers to questions in Visual Genome could be answered through a combination of words included in the ground truth scene graphs (excluding a 7% of questions involving counting from this figure). However, after using the textual labels of all scene graphs to query DBpedia, WebChild, and ConceptNet, nearly twice as many questions (79.58%) could be correctly answered. These results show that the type of information residing in common and commonsense general-purpose knowledge bases is complementary to the semantic annotations provided with Visual Genome. Thus, there is the potential for augmenting datasets developed for benchmarking on visual tasks with knowledge coming from other external sources.

Another relatively less explored dataset we identified is **ShapeNet** (Chang *et al.*, 2015), a large-scale collection of 3D object models. ShapeNet is split into ShapeNetCore and ShapeNetSem. Albeit including a lower number of models than ShapeNetCore, ShapeNetSem is augmented with richer annotations describing the physical properties of objects, e.g., absolute size estimations, upright and front orientation and others (Savva *et al*., 2015). Most recently, "part of" annotations for a subset of ShapeNet models spanning across 24 object categories were released as **PartNet** (Mo *et al.*, 2019).

## 3. The ingredients of Visual Intelligence

### 3.1. A top-down approach

Lake *et al*. (2017) have recently suggested a set of *core ingredients* that characterize the way we think and learn. Their discussion broadly concerns human intelligence as a whole and impacts all sub-fields of AI. From Lake *et al*., (2017) we borrow those ingredients which are relevant to the task of Image Understanding, namely *"learning as model building"*, *"intuitive physics"* and *"thinking fast"* (renamed *fast perception* in the following). We further complement these ingredients with other principles characterizing humans' Visual Intelligence, based on Donald Hoffman's seminal book, "Visual Intelligence: how we create what we see" (2000). These additional ingredients are *"spontaneous morphing"*, *"generic views"*, and *"motion vision"*. We use the identified principles as a reference framework to discuss other relevant theories of AI and Visual Cognition.

---

[2] https://verbs.colorado.edu/verb-index/vn3.3

Each Visual Intelligence component brings along two levels of requirements: (i) a set of required reasoning capabilities, and (ii) a set of knowledge requirements. Both sets of requirements are listed at the end of each paragraph.

**Learning as model building.** Humans can recognize the boundaries between different physical entities (Rosch *et al.*, 1976; Hayes, 1988), the natural structure behind each observation (Minsky, 2007) and discern what is relevant from what is irrelevant (Rosch *et al.*, 1976, Brooks, 1991). Thanks to these perceptual abilities, they can build a fine-grained mental model of their environment (Lake *et al.*, 2017; Pearl, 2018), where new concepts can be formed, by combining previously learned concepts (Chomsky, 2010). The latter capability has been also referred to as **learning-to-learn** (Lake et al., 2017) or **meta-learning** (Chen and Liu, 2018).

Moreover, humans can observe the causes that generated a specific concept. This **causal knowledge** leads to learning more robust concepts, which can be reused flexibly in different scenarios and expanded to accommodate new concepts (Davis and Marcus, 2015). Conversely, state-of-the-art Machine Learning models can only find strong correlations, i.e., recognize patterns, in the provided input dataset. Back to our example, HanS would need to know that electric devices can cause hot surfaces and that the proximity to paper can spark a fire, regardless of how often this event has ever occurred. Similarly, commonsense allows humans to discern between (frequent) correlations and causality and therefore to handle anomalies and infrequent events way more effectively than machines do. **Long-tail phenomena** (Davis and Marcus, 2015) are in fact particularly difficult for pattern recognition algorithms to detect, precisely because they are rarely observed.

This component thus requires ▶ *incremental object learning* and ▶ *causal reasoning* capabilities. It also requires: ▶ *higher-level object representations* that can be expanded opportunistically, as new concepts are learned; ▶ *hierarchical object taxonomies* where new concepts are represented as a combination of existing concepts; ▶ *cause-effect relations* between concepts (*including infrequent ones*).

**Intuitive physics.** One of the causal world models which children excel at constructing since their very first months is that which adheres to intuitive principles of physics such as solidity, continuity, inertia, and others (Spelke *et al.*, 1995). Since Hayes' "Naïve Physics Manifesto" was published in 1978, many have advocated the need to integrate intuitive physics, or commonsense physical knowledge in AI systems (Davis and Marcus, 2015; Lake *et al.*, 2017). In Hayes' view, these commonsense, physical properties of objects (e.g., shape, orientation, physical states, forces) are organized as clusters, i.e., neighbourhoods of concepts, tightly related through several axioms (Hayes, 1988). In this sense, intuitive physical properties also play a role in how we categorize objects. For instance, our priors about the typical relative sizes of objects strongly influence the way we interpret perspective in images. If we were shown a picture depicting a very large cup and a relatively smaller (but similarly shaped) rubbish bin, we would still be able to disambiguate the two. We would conclude that the cup is in the foreground and that the bin is in the background, because we know that cups are typically smaller than bins.

In sum, the envisaged system would need to include ▶ *a physics reasoner*, embedding ▶ prior knowledge of *physical properties of objects*, such as size, natural orientation, weight, surfaces that typically support other objects, and others.

**Spontaneous morphing.** Another ingredient that makes human Visual Learning so efficient is what Hoffman (2000) defines as spontaneous morphing and Lake *et al.* (2017) call **compositionality**. In other words, humans process visual concepts as a combination of parts and relations between these parts (Lake *et al.*, 2017).

Although space is continuous, humans discretize it to make decisions in a timely and efficient manner, i.e., by means of **qualitative spatial reasoning**. Spatial primitives like containment or contact, help children learn to differentiate between the self and the surroundings and to categorize the world as a collection of "things" (Piaget, 1956), even before knowing what these things are (Hoffman, 2000; Rosch *et al.*, 1976). Dividing objects into parts and spatially relating these parts with one another is essential to object recognition. We rarely see objects in their entirety and, as we move, different parts become visible and other disappear from our visual field. Moreover, many objects include movable parts, which can be configured differently. Therefore, identifying the distinct parts of objects allows us to recognize them robustly, from different viewpoints and under different configurations. To adhere to the principle of spontaneous morphing, a desirable system should include ▶ a *fine-grained segmentation module* to recognize the object sub-parts; and ▶ *geometric* and ▶ *spatial reasoning* capabilities. The types of knowledge properties which can support these capabilities are the typical ▶ *part-whole relations* (forming a *partonomy*) and ▶ *Qualitative Spatial Relations (QSR)* between objects.

**Generic views: how we construct depth.** The images cast at the back of our eyes, i.e., retinal images, are 2-dimensional. We construct their representation in the 3D world mentally, thanks to our Visual Intelligence (Hoffman, 2000). Specifically, we construct only those 3D models for which the retinal image provides a generic (i.e., stable) view. As a result, there exists a set of preferred, or canonical, views from which we can recognize objects more rapidly and effectively. These 2D views have certain shape and colour attributes.

First, the way we typically draw contours on 2D images and segment the objects is non-arbitrary. Hoffman (2000) refers to the 2D shapes we construct as **"subjective"** because our mind constructs them. Yet, these shapes are also universal, because we all construct them according to the same rules. This view aligns with findings that showed that humans categorize object by similarity to **prototypical shapes**, obtained from averaging all contours of objects belonging to a certain class (Rosch *et al.*, 1976; Rosch, 1999).

Moreover, we interpret our retinal images based on the

**colours** we construct. Colour, however, causes a lot of ambiguity. For instance, shapes printed in the same ink on a sheet of paper can look different from one another under different light sources or depending on which other colours are occupying our visual field (Hoffman, 2000). To overcome this ambiguity, the human mind poses specific constraints on the way we decide to apply colours to both the objects and the light sources we observe. Although the mechanisms that allow humans to approximate similar colours despite changes in lighting are still unclear, there is evidence that our perception of colours is based on principles of stability. When interpreting retinal images, we select the most stable combination of shape, colour and light, i.e., the one for which perturbations in shape, colour and luminance result in the smallest changes to the image.

This evidence suggests that ▶*classifying objects based on their visual similarity to generic (or prototypical) 2D views* is another important capability for Visual Intelligence. This capability also implies to have access to ▶ a set of *generic 2D views of objects*, from which one can more easily extract *prototypical shapes* and *stable colour regions*.

**Motion Vision.** Motion is also constructed by our Visual Intelligence (Hoffman, 2000), and plays a role in object categorisation. Eleanor Rosch was the first one to show, through an extensive series of experiments, that we group objects into *basic categories* not only based on their shared attributes and shape similarity, but also based on **common motor programs**, i.e., sequences of human motor actions used to interact with these objects (Rosch *et al*., 1976). Gibson (1979) later called these properties, or typical uses of objects, *affordances*.

The **motion trajectories** of objects also aid their categorisation and long-term representation. Cognitive studies (Kourtzi and Nakayama, 2002; Wallis, 2002) have suggested that the human brain maintains two distinct representations (or *signatures*) for static and moving objects. While static objects are represented by combining a set of "known" static views "within a limited spatial range", moving objects are represented via the interpolation of sequences of images along the object motion paths, "even for long motion trajectories" (Kourtzi and Nakayama, 2002).

Therefore, ▶*object tracking* and ▶ *action recognition* across temporally ordered frames are two other required capabilities. These two components can benefit from the integration of prior knowledge of ▶ the *typical affordances* and ▶ *motion trajectories* of these objects.

**Fast perception.** Humans learn to recognize unknown objects very rapidly, often from the very first exposure i.e., one-shot learning (Lake *et al*., 2017). Thus, another requirement for systems to exhibit Visual Intelligence and rapidly adapt to changes in the environment, is to maximize their inference speed. There is evidence that inference times are higher when querying external repositories, especially when computationally expensive physics game engines are involved (Beetz *et al*., 2018), than when applying off-the-shelf Deep NN-based methods (Lake *et al*., 2017). Thus, a promising direction is capitalizing on Deep Learning methods which have ensured near real-time object recognition performance on known categories (Redmon and Farhadi, 2018; Ren *et al*., 2015), and combine those with properties retrieved from external knowledge bases (Lake *et al*., 2017; Aditya *et al*., 2019; Gouidis *et al*., 2019).

### 3.2. A bottom-up approach

Having defined a set of ingredients for Visual Intelligence in a top-down fashion, we can now use them to frame a qualitative analysis of the classification errors encountered in our object recognition trials, which have been carried out by means of a purely ML-based method. To this purpose, we relied on a two-branch Network with a ResNet50 backbone, which was pre-trained on ImageNet, and applied weight imprinting to the softmax classification layer (Chiatti *et al*., 2020). We fine-tuned the NN across 25 object classes, five of which are specific to Health and Safety monitoring, as more thoroughly described in (Chiatti *et al*., 2019). We collected 295 test images (worth 896 distinct object regions) at the Knowledge Media Institute (KMi), using a Turtlebot mounting an Orbbec Astra Pro RGB-D monocular camera. Frames were collected in a temporal sequence, during one of HanS' patrolling rounds, and stored at their maximum resolution (i.e., 1280x720). These data were not re-sampled and class cardinalities are representative of the natural occurrence of objects along the scouting route: e.g., HanS is more likely to spot fire extinguishers than windows. To ensure focus on classification errors and leave out segmentation errors, object regions were annotated manually. From the 896 original regions, we exclude 35 regions with ambiguous annotations, i.e., where the annotated rectangular region encloses more than one object. For example, the algorithm mistook a region labelled as plant for a coat stand when the bounding box enclosed both a plant and part of a coat stand. 272 (31.59%) of the remaining object regions were misclassified and form the basis of our error analysis.

**Qualitative error analysis.** We annotate the classification errors in each test image as distinct rows in a Boolean matrix, as shown in Table 1. Columns in the matrix are the missing capabilities or knowledge properties which would have helped: (i) to identify the ground truth object, or (ii) to rule out the incorrect object. For instance, if a bin was mistaken for a cup twice, we will use two distinct rows, because the models of objects and circumstances depicted in each image can be different. Let us imagine that, in this example, only the second row depicts a recycling bin, with a visible sign stating: "general waste". In both cases, by knowing that a paper bin and a mug, regardless of their similar shape, have significantly different sizes, HanS would not confuse them.
Other intuitive physics properties (e.g., natural orientation, or solidity) are not helpful in this case, because both items are more developed vertically than horizontally and both are containers. Moreover, by observing that the object to classify is lying on the floor, one can conclude that mugs are not a likely candidate. Thus, the robot's spatial reasoning capabilities and prior knowledge are relevant to both cases.

| Ground truth class | Predicted class | Intuitive Physics ||| Spatial Reasoning | Machine Reading |
|---|---|---|---|---|---|---|
| | | Size | Ori. | Solid. | | |
| Bin | Mug | T | F | F | T | F |
| Bin | Mug | T | F | F | T | T |

Table 1 Example of Boolean matrix used for analyzing errors.

However, the capability of reading the words "general waste" applies only to the case of the recycling bin, i.e., to the second row in Table 1. Since the capability of reading signs is not included in the set of top-down requirements generated by the analysis in Section 3.1, we need to add an additional column to the matrix. Finally, for each column, we count the number of rows where it was marked as relevant (e.g., size impacted 2 out of 272 cases). The resulting error counts, aggregated by component type, are reported in Figure 1.

As shown in Figure 1, our analysis demonstrates that, with the partial exception of model building, all other components of the proposed framework for Visual Intelligence play a very significant role in object recognition. Their integration within a Visual Intelligence architecture for HanS has thus the potential to significantly improve its performance. In what follows, we analyze the links that have emerged between errors and epistemic components in detail.

**Model building.** Overall, the only type of causal knowledge and causal reasoning which was found to be relevant for mitigating the object recognition errors has to do with Intuitive Physics. Indeed, other types of causal relations, albeit still essential to Image Understanding (e.g., the proximity of an electric heater to a pile of paper is likely to cause a fire), apply to the visual inference steps following the object recognition phase. In 5.15% of cases, the main cause of error is the inadequacy of the object taxonomy chosen for these trials. For instance, 8.33% of books were classified as paper, 7.14% of bottles were mistaken for mugs, i.e., a semantically similar class. With access to a hierarchical taxonomy of concepts (which is another requirement of model building), classification can be tackled in steps and more conservatively, e.g., by first recognizing that the object is a drink container and then focusing on whether it is a bottle or a mug. This requirement also relates to incremental object learning and meta-learning. Namely, more accurate predictions could be made on new object types, i.e., unseen at training time, by analogy with other semantically related concepts. For example, the KMi foosball table was not part of the 25 training classes; hence, it was not recognized in the test set. However, if HanS recognized it as a desk and knew that desk and table are synonyms, and that foosball table is a special type of table, it would be on the right path to recognize this novel object. Therefore, the overall impact of the model building requirement goes beyond the numbers reported in Figure 1, which are only based on known object regions.

**Intuitive Physics.** The capability to reason about the physical properties of objects was found to be the most impactful component across all error cases. Specifically, we identified three main components of Intuitive Physics which were crucial for correction: (i) the objects' *relative size*s (in 73.53% of cases), (ii) *solidity* qualities, i.e., concave, or "container-like" solids, as opposed to convex and saddle solids (in 40.07% of cases), and (iii) their *natural orientation*, e.g., coat stands are typically oriented in an upright position (in 25.74% of cases). For instance, 23.08% of armchairs were mistaken for paper bins despite their difference in size; 36.11% of desks were mistaken for coat stands, even though the width of desks is normally greater than their height, whereas the height of coat stands is normally greater than their width; 9.52% of bottles, i.e., containers, were confused for radiators. However, the object natural orientation was misleading, in some cases: 11.54% of plants were mistaken for coat stands.

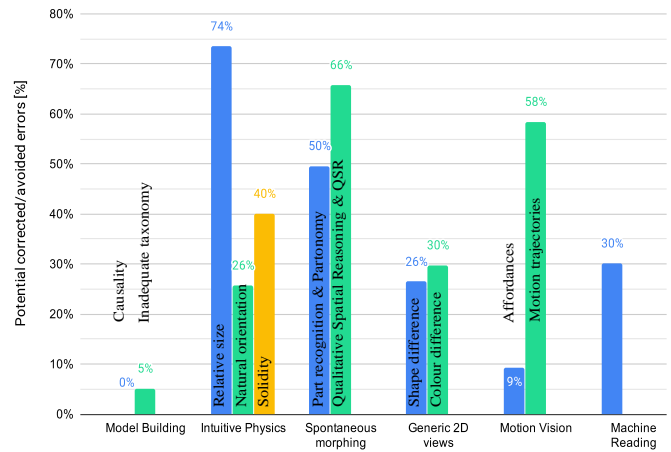

Figure 1: Percentage of cases where a specific component of Visual Intelligence would help correcting or avoiding the classification error.

**Spontaneous morphing.** This analysis also confirmed the importance of *spatial reasoning capabilities and QSR*, which impacted 65.81% of cases. For instance, 13.57% of fire extinguisher signs were mistaken for a desktop PC. However, knowing that the recognized rectangular shape is hanging on a wall, above a fire extinguisher, would have significantly scoped down the possible predictions. Similarly, 50% of monitors were misclassified as radiators, even though radiators, usually, are not laying on top of a desk.

The second component of the spontaneous morphing ingredient, i.e., *the capability to recognize the different visible parts composing an object*, was found relevant to 49.63% of recognition cases. For instance, 22.54% of doors were classified as rubbish bins and 21.13% as boxes. However, KMi doors have distinctive, visible sub-parts which differentiate them from bins or boxes, such as door handles and glass panels. Thus, the access to a *partonomy* detailing the components of a door would help in this case.

**Motion Vision.** Because these data were collected in temporal order, Motion Vision was found to be another important component. The capability to track objects across successive frames and the prior knowledge of static/moving objects (i.e., *motion trajectories*) were found to impact 58.46% of cases. For instance, 7.53% of people walking by were misclassified as coat stands (i.e., static objects) in specific frames. 9.09% of radiators, which, on the contrary, are very unlikely to change their position, were not recognized consistently across successive frames. Only 38 out of the 295 test images depict human interactions with objects. As a result, the knowledge of common *object affordances* was found relevant to only 9.19% of error cases. However, if we only consider the subset images representing human interactions with objects, we find that object affordances would have helped correcting 57.89% of these images. For example, recognizing a person who is staring at an unspecified object, while leaning over a desk and holding a mouse, is a strong cue that the object is a monitor.

**Generic 2D views.** In 26.47% of cases, objects were mistaken for other classes, irrespective of the fact that the two classes exhibited highly different *shapes*, or 2D contours (e.g., bottles classified as radiators). Similarly, in 29.78% of cases, objects were confused for one another despite their clear-cut *colour* differences. For instance, a red fire extinguisher sign was classified as an emergency exit sign, even though all emergency exit signs in KMi are green. Interestingly, the shape and colour similarity between 2D views of different object classes led to recognition errors, in some cases. For instance, 21.13% of doors were confused for boxes likely due to their rectangular silhouette; 5.56% of fire extinguishers were classified as bottles; a green bottle was mistaken for an emergency exit sign; three windows with white blinds were classified as radiators. This evidence can explain why the requirement of generic 2D views, overall, impacted a relatively lower percentage of error cases than other ingredients, similarly to the case of natural orientation (Figure 1).

**Machine Reading.** We identified one additional cause of error which could not be mapped to the other top-down components: ▶ the lack of *Machine Reading capabilities* (30.15% of cases). This requirement was found to be particularly relevant to the case of labelled items and signs, which appear frequently in the domain of interest. Recycling bins, for instance, are explicitly signaled with cue words such as "general waste", "cans & bottles" and so forth. Similarly, fire extinguisher signs include standard terms such as "carbon dioxide" or "water". Thus, the capability to not only recognize the characters appearing in an image, but also understand their meaning (i.e., going from Optical Character Recognition to Machine Reading), would significantly aid the recognition of signs and labelled items in KMi.

## 4. Knowledge Base Evaluation

Based on the identified requirements, the next step is to assess to which extent the current knowledge bases can provide the missing knowledge properties. Therefore, in what follows, we focus on the types of knowledge identified in Section 3 (also listed in the heading of Table 2), leaving out the required reasoning capabilities. Specifically, the bottom-up error analysis highlighted the need for a Machine Reading component, in addition to the top-down requirements. However, because this component is a missing capability, rather than a type of knowledge, it is omitted from this evaluation. As shown in Table 2, none of the epistemic requirements of Visual Intelligence is fully met by state-of-the-art KBs. In the remainder of this Section, we discuss the level of coverage available for each component, as well as the identified gaps and limitations.

**Hierarchical taxonomy.** To assess whether the selected KBs can adequately represent newly learned concepts, as a combination of known concepts, we indicate if they adhere to a hierarchical taxonomy. As shown in the left-most column of Table 2, the majority of the selected KBs already provide links to WordNet, hence this is a natural choice to play the role of a reference taxonomy.

**Cause-effect relations.** Model building is also supported by equipping robots with prior knowledge of both frequent and infrequent cause-effect relations. Probase is the KB, among the reviewed ones, which provides the largest set of long-tail relations (e.g., cockroach is a revolting animal, trash can cleaning is the yuckiest cleaning job, and others). In particular, the type of causal relations of interest involves everyday objects (e.g., heater is a heat source), and indeed Probase, ConceptNet and ASER provide this type of cause-effect relations, at least for a subset of objects. Among these, Probase, in particular, covers the largest portion of specialized terms which are relevant to our application (e.g., fire extinguisher sign is a fire safety sign/equipment).

ATOMIC represents events in an agent-centric rather than object-centric way and is more focused on the abstract and social causes and effects of certain events (e.g. Person X leaves object on the table and feels forgetful as a result, but without emotionally affecting others or causing them to want to do anything). Therefore, ATOMIC covers only a subset of the causal relations of interest. Similarly, KnowRob specializes on the observed outcomes of specific manipulation actions (e.g., setting up a table). However, the set of events of interest spans beyond object manipulation demonstrations.

Overall, it is more difficult to organize coherently and reuse effectively the causal knowledge residing in KBs which mix different knowledge types (Probase and ConceptNet), compared to KBs specialized on causality (ASER and ATOMIC). All relations in Probase are generically IsA relations and ConceptNet uses several different relation types to entail causality (e.g, HasPrerequisite, Causes, HasSubevent and others). Moreover, Probase, despite its broader coverage, does not adhere to a standardized, hierarchical taxonomy. One way to overcome this limitation would be to map a subset of concepts in Probase to WordNet. In this way, the causal verb groupings in UVI could be used to link causally related concepts. For instance, the verb *to move*, which, in Probase, is linked to concepts such as *event* and *manipulation instruction*, is grouped together, in UVI, with properties such as *to cause motion* or *to change position on a scale*.

**Intuitive Physics Knowledge.** The bottom-up analysis presented in Section 3.2 highlighted three types of physical properties which are crucial for robots to improve their capability to recognize objects. These are, in descending order of impact: (i) the *relative size* of objects, (ii) their *solidity qualities*, and (iii) their *natural orientation*. The first property is provided, for a subset of objects, in KnowRob, Wikidata and ShapeNet. Among these, ShapeNet is the KB which covers the highest number of object categories of interest. Indeed, the KnowRob physics engine was tailored to a specific environment and object catalogue (e.g., the kitchen utensils needed to make a pizza). Properties in Wikidata are even more scarce and scattered, because Wikipedia infoboxes follow varying templates, based on the object being described. As a result, certain furniture pieces, e.g., chairs, are related to their real-world dimensions (height, width, depth), but the same properties are not available for other relevant items, e.g., fire extinguishers.

| KB | Knowledge Requirements of Visual Intelligence | | | | | | | | Accessibility |
|---|---|---|---|---|---|---|---|---|---|
| | Hierarchical Taxonomy (linked to WordNet?) | Cause-effect relations | Intuitive Physics Knowledge | Part-whole relations | QSR | Generic 2D views | Object affordances | Motion trajectories | |
| Unified Verb Index (UVI) | yes | ◉ | | | | | ◉ | | High |
| KnowRob/Open-EASE | yes | ◉ | ◉◉ | ◉ | | | ◉ | ◉ | Partial |
| DBpedia | yes | | ◉ | ◉ | ◉ | | ◉ | | High |
| Wikidata | yes | | ◉ | ◉ | ◉ | ◉ | ◉ | | High |
| Probase | no | ◉◉ | | ◉ | ◉ | | | | Partial |
| NELL | no | | | | ◉ | | ◉◉ | | Adequate |
| ConceptNet | yes | ◉◉ | ◉◉ | ◉ | ◉ | | ◉ | | High |
| WebChild | yes | | ◉ | ◉ | | | ◉◉ | | High |
| ATOMIC | no | ◉ | | | | | | | High |
| ASER | no | ◉◉ | | | | | ◉ | | Adequate |
| Visual Genome (VG) | yes | | ◉◉ | ◉◉ | ◉◉ | ◉ | ◉◉ | | High |
| ShapeNet/ PartNet | yes | | ◉◉ | ◉◉ | | ◉◉ | | | Partial |

Table 2: Summary of the KB evaluation. The level of coverage of each knowledge requirement is marked by using from one to three dots.

The second property, which concerns the solid surfaces of objects, can be derived from simulated 3D models, as in the case of KnowRob and ShapeNet, or from more explicit textual descriptions (e.g., desk is a flat horizontal surface, bottle is a container), as in the case of Probase, NELL, DBpedia and Wikidata.

Third, the natural orientation of objects is also embedded in the KnowRob simulation engine. However, ShapeNet covers a larger subset of objects of interest than KnowRob, and explicitly annotates certain 3D models as upward and front oriented.

Besides the physical properties directly impacting object recognition, in order to interpret the current level of risk, HanS also needs to know: (iv) the component materials of objects (e.g., book is made of paper), (v) the physical properties of these materials (e.g., paper is flammable), and (vi) how these properties compare to one another (e.g., paper is more flammable than other materials). Textual descriptions of the object fabrication materials can be found in DBpedia, Probase, NELL, ConceptNet and WebChild. However, only ShapeNet, VG and Wikidata accompany these annotations with visual examples. In ShapeNet, the ratios of component materials are aggregated by class, rather than being annotated for each object model. The resulting material compositions are noisy, especially for object classes which comprise of several different models (e.g., chairs, desks). VG, instead, provides a more reliable alternative, because assertions such as *chair is wooden* are grounded to the specific chair instance depicted in the observed image. Moreover, the variety of covered objects and object models is greater in VG than in Wikidata. Nonetheless, Wikidata, through WordNet, can provide a link to other physical properties of interest (e.g., paper is flammable) which are available in DBpedia, Probase, ConceptNet and WebChild. Crucially, ConceptNet and WebChild also represent material properties in comparative terms – e.g., paper is easier to burn than wood. However, WebChild was found to be highly unreliable. For instance, *paper* is considered *a substance of the Internet*, *bicycle* is *physically smaller*, but also *more abundant* than *car*.

**Part-whole Relations.** For a subset of objects, part-whole relations are provided in VG and PartNet (e.g., white door with silver knob, refrigerator has power cord). Both KBs also provide the annotated image regions depicting these relations. The object masks used for annotation in PartNet are more accurate than the rectangular regions in VG, however the latter one covers a larger set of objects.

For the other knowledge bases which embed partonomies of concepts, the limitations discussed in the context of causal and intuitive physics knowledge also apply in this case. First, DBpedia and Wikidata include less part-whole relations of interest due to the highly variable Wikipedia templates. Similarly, compositional descriptions in NELL are purely textual and unstructured (e.g., office chair *has* armrests, back rest *of* office chair). Third, Probase does not adhere to a coherent partonomy. Similarly, part-whole relations in ConceptNet are spread across different relation types, e.g., ThingsLocatedAt, ThingsWith and others. Lastly, WebChild provides noisy assertions – e.g., *lake* is part of *paper*; *humans* have *snouts*.

**Qualitative Spatial Relations (QSR).** Another important requirement we identified is the capability to single out predictions which appear in atypical locations (e.g., a radiator on top of a desk) and are thus more likely to be incorrect. In other words, we are looking for (i) QSR between objects appearing in the same image and for (ii) ways to measure the typicality of these QSR. The QSR provided in VG (e.g., fire extinguisher ON wall, radiator ON floor) meet both requirements. First, VG provides object-object relations represented at the image level, whereas the spatial relations provided with the other KBs highlighted in column 5 of Table 2, either mix different levels of granularity (e.g., object-object with object-room relations) or are completely unstructured (e.g., computer is often found in an office). This issue is particularly pronounced in the case of ConceptNet and WebChild, where QSR are mixed with part-whole relations (e.g, *CPU* is a thing located at *computer*; *radiator* is in spatial proximity with *water*, *air* and *bathroom*). Second, since spatial relations in VG are annotated for each image, their frequency of occurrence throughout the collection can provide a measure of their typicality (Chiatti *et al.*, 2019).

Nonetheless, all the reviewed KBs are missing some of the spatial relations which are very specific to our use case scenario (e.g., that fire extinguisher signs are hanging on the wall, right above fire extinguishers – see also Section 3.2).

**Generic 2D views.** Another important requirement of Visual Intelligence is the access to *generic 2D views*, to extract the *prototypical shapes* and *stable colour regions* characterizing different objects. VG, on the one hand, provides, for a subset of object regions, annotations of their shape, colour and texture (e.g., fire extinguisher is red, trash can is round).

Compared to the natural scenes in VG, 2D object models in ShapeNet are pre-segmented and simplified (i.e., synthetically generated), allowing to control for background noise and occlusion (an issue which also affected the manually annotated regions in our real-world dataset - Section 3.2).

Wikidata also provides a set of exemplary images for each one of its entries. However, VG and ShapeNet offer significantly larger image collections, ranging across different object models.

**Object affordances.** Observing the different uses and motor actions associated with objects can also aid their recognition. Thus, HanS would need access to the human interactions with those objects and H&S equipment which are commonly found in an office space. These interactions are varied and not only concerned with the active manipulation of objects, i.e., the main focus of KnowRob. For instance, knowing that a person is staring at a rectangular object while leaning on a desk would help classifying the object as a screen.

Other KBs provide a broader set of the affordances of interest, as shown in Table 2. However, each one of these resources comes with its limitations.

Descriptions in NELL, DBpedia and Wikidata are purely textual and unstructured (e.g., I stood up from office chair; chairs are commonly used to seat a single person). Similarly, ATOMIC and ASER include descriptions of the actions that occurred during a certain event, but the type of representation used, in both cases, is conceived to express cause-effect relations. As a result, extracting action sets from these representations would be an expensive and error-prone process.

Conversely, ConceptNet, WebChild and VG represent actions into more structured predicates: "UsedFor" and "CapableOf" (ConceptNet); "activity" relations (WebChild); action predicates (VG - e.g., woman pouring water). Unfortunately, both ConceptNet and WebChild include ambiguous affordances, mixing different word senses. For instance, *monitor* can be both a type of *input device* and a *supervisor* and is thus associated to the activity *become monitor*. On the contrary, while VG includes a more limited set of predicates, these are canonicalized with respect to the WordNet taxonomy. Thus, they are both more coherent and can also be mapped to other linguistic resources, e.g., UVI.

**Motion trajectories.** Notably, none of the reviewed KBs explicitly encodes the common *motion trajectories* related to objects, to categorize them as static (e.g., a radiator), movable (e.g., a water bottle) or moving (e.g., a person). In principle, the episodic memories in KnowRob would allow to infer these motion trajectories, because the observed actions are annotated at specific timestamps. However, as already mentioned, these episodic memories are constrained to specific use case scenarios.

**Pragmatics.** In Table 2, we also report the level of accessibility of each KB on a qualitative scale. Accessibility is judged as "Partial" in cases where only part of the encoded knowledge is openly available, e.g., KnowRob. Similarly, to the best of our knowledge, the latest release of ShapeNetSem provides annotations for only a subset of the physical properties described in (Savva *et al.*, 2015); Probase is just a portion of the MS Concept Graph. All the remaining KBs are considered to provide an adequate or even high level of accessibility, depending on whether they also provide an intuitive browser and API services.

## 5. Conclusion

Despite the recent popularity of Computer Vision methods based on Deep Neural Networks, machine Visual Intelligence is still inferior to human Visual Intelligence in many ways. Inspired by this evidence, and by related works in AI and Visual Cognition, we have identified a set of epistemic requirements to equip systems with more powerful Visual Intelligence capabilities. Since our focus is on service robotics, we therefore grounded a set of core theoretical ingredients into a concrete, real-world scenario. Through this combination of top-down and bottom-up components, we shed a light on the required set of capabilities and knowledge properties for service robots to exhibit human-like Visual Intelligence. As such, the findings presented in this paper provide a reference framework for choosing which components to prioritize and leverage in the development of knowledge-enriched vision systems for service robots.

Moreover, we examined the extent to which state-of-the-art knowledge bases can support the knowledge requirements highlighted by this framework. Crucially, we found that none of the reviewed KBs meets these requirements in full. The three most impactful knowledge attributes exposed by the bottom-up analysis (the object relative sizes, QSR and typical motion trajectories) are covered only for a limited set of objects. In particular, a major limitation concerns the lack of knowledge representations to categorize objects based on their motion trajectories, e.g., as static or moving.

The identified gaps serve as a research agenda for the development of improved knowledge representations.

In particular, in our future work, we will concentrate on capitalizing on ShapeNet, Visual Genome and ConceptNet to develop a more robust intuitive physics engine, with the aim of achieving a significant improvement with respect to HanS' Visual Intelligence capabilities.